\documentclass{article}
\usepackage{spconf,amsmath,graphicx,booktabs}
\usepackage{amssymb}
\usepackage[pagebackref=true,breaklinks=true,letterpaper=true,colorlinks,bookmarks=false]{hyperref}
\usepackage{adjustbox}
\usepackage{xcolor}
\usepackage{float}
\usepackage{fancyhdr}

\usepackage[linesnumbered,ruled,boxed,norelsize,vlined,commentsnumbered]{algorithm2e}

\usepackage{array} 

\title{Top-K POOLING WITH PATCH CONTRASTIVE LEARNING FOR WEAKLY-SUPERVISED SEMANTIC SEGMENTATION}
\name{Wangyu Wu$^{1,2}$, Tianhong Dai$^{3}$, Xiaowei Huang$^{2}$, Fei Ma$^{1^{*}}$, Jimin Xiao$^{1^{*}}$\thanks{$^{*}$Corresponding authors}}
\address{$^{1}$Xi’an Jiaotong-Liverpool University\quad
$^{2}$The University of Liverpool \quad $^{3}$University of Aberdeen
}

\begin{document}
\maketitle

\setlength{\footskip}{15pt} 
\begin{abstract}
Weakly Supervised Semantic Segmentation (WSSS) using only image-level labels has gained significant attention due to cost-effectiveness. Recently, Vision Transformer (ViT) based methods without class activation map (CAM) have shown greater capability in generating reliable pseudo labels than previous methods using CAM. However, the current ViT-based methods utilize max pooling to select the patch with the highest prediction score to map the patch-level classification to the image-level one, which may affect the quality of pseudo labels due to the inaccurate classification of the patches. In this paper, we introduce a novel ViT-based WSSS method named \textit{top-K pooling with patch contrastive learning} (TKP-PCL), which employs a top-K pooling layer to alleviate the limitations of previous max pooling selection. A patch contrastive error (PCE) is also proposed to enhance the patch embeddings to further improve the final results. The experimental results show that our approach is very efficient and outperforms other state-of-the-art WSSS methods on the PASCAL VOC 2012 and MS COCO 2014 dataset.
\end{abstract}
\begin{keywords}
weakly-supervised semantic segmentation, vision transformer, top-K pooling, contrastive learning
\end{keywords}
\section{Introduction}
\label{sec:intro}
Weakly Supervised Semantic Segmentation (WSSS) is an evolving approach in the field of computer vision, which aims to generate pixel-level labels by utilizing weak supervision signals to heavily reduce the cost of annotations. Among the various weak supervision signals, the image-level label is the easiest to obtain. However, it only provides limited information for semantic segmentation tasks.

Previous  WSSS approaches that rely on image-level class labels typically predict pseudo labels using class activation maps (CAM)~\cite{zhou2016learning}. Nonetheless, CAMs exhibit limitations in accurately estimating both the shape and localization of classes of interest objects~\cite{chen2022class}, motivating researchers to employ additional refinements between the initial pseudo labels and the final pseudo labels generation. These refinements often involve the multi-stages architecture, as observed in PAMR~\cite{araslanov2020single}, which further increases the complexity. More notable refinement strategies haven been described in IRNet ~\cite{ahn2019weakly} and AdvCAM~\cite{lee2021anti}. In recent years, due to the limitations of CAM, researchers have turned to leverage ViT-based~\cite{dosovitskiy2020image} frameworks for WSSS~\cite{rossetti2022max,ru2023token,xu2022multi,ru2022learning}. ViT-PCM~\cite{rossetti2022max} uses patch embeddings to infer the probability of pixel-level labels. Methods~\cite{ru2023token,xu2022multi} utilize ViT to substitute CNN, thereby enhancing the capability of CAM in object recognition. AFA~\cite{ru2022learning} suggests acquiring reliable semantic affinity through attention blocks, aiming to enhance the initial coarse labels. However, the current ViT-based methods utilize the global max pooling to select the patch with the highest prediction score to project the patch-level classification to the image-level one, which may affect the final performance due to misclassifiaction.

\begin{figure}[t]
\centering
\includegraphics[width=0.7\linewidth]{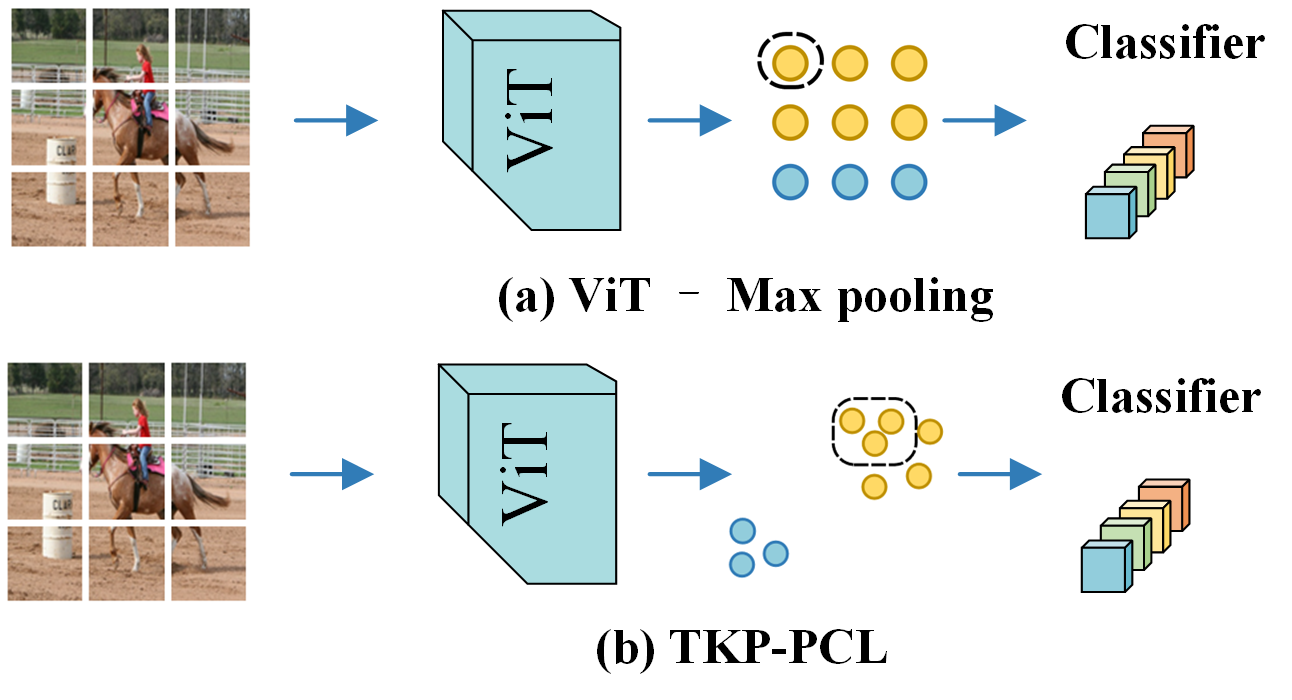}
\vspace{-0.3cm} 
\caption{(a) The previous ViT-based method~\cite{dosovitskiy2020image} only projects the class prediction of a single patch of the highest prediction score into the image-level classification. (b) Our TKP-PCL projects the class prediction from top-K patches into the image-level classification and use contrastive loss to enhance patch embeddings.}
\label{fig:idea}
\vspace{-0.3cm} 
\end{figure}


The major challenge of WSSS is to improve the accuracy of the initial pseudo labels based on image-level labels. To this end, we propose an effective approach called \textit{top-K pooling with patch contrastive learning} (TKP-PCL, see Fig.~\ref{fig:idea}) and our main contributions can be summarized into threefold:

1) We propose a novel ViT-based framework that enhances pseudo label prediction accuracy by employing top-K pooling to address the limitations caused by misclassified patches.

2) A patch contrastive error (PCE) is proposed to improve intra-class compactness and inter-class separability of patch embeddings to further improve the quality of pseudo labels.

3) In the experiments, our proposed approach outperforms other state-of-the-art methods for the segmentation task on the PASCAL VOC 2012~\cite{everingham2010pascal} and MS COCO 2014 dataset~\cite{lin2014microsoft}.

\begin{figure*}[h] 
\begin{center}
    \includegraphics[width=0.8\linewidth]{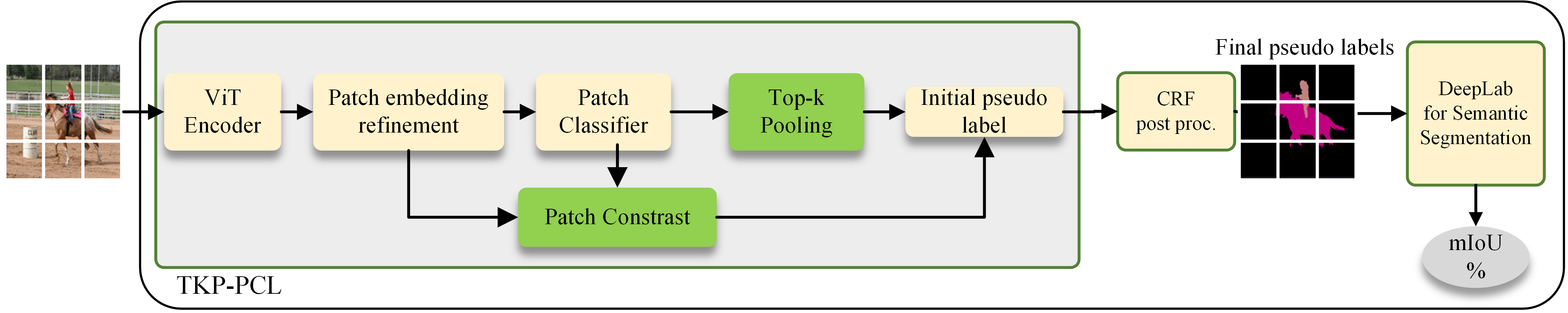}
    \vskip -0.3cm
   \caption{The illustration of the main components of the framework and how TKP-PCL method learns to obtain the initial pseudo labels. Subsequently, the initial pseudo labels are refined using a Conditional Random Fields (CRF)~\cite{krahenbuhl2011efficient} and is then passed to the a selected semantic segmentation model (i.e., DeepLab~\cite{chen2017deeplab}). The green blocks indicates the primary contributions of our method.}
    \label{fig:framework}
\end{center}
\end{figure*}

\section{RELATED WORKS}
\subsection{WSSS with Image-level Labels.}
Existing WSSS methods commonly rely on image-level class labels as the cheapest form of supervision. Approaches using image-level class labels have traditionally been based on CAM methods~\cite{zhou2016learning}, employing a standard multi-label classification network. The CAMs are derived by applying global average pooling (GAP) to the feature maps of the last layer, followed by concatenation into a weights vector. This vector is then connected to the class prediction through Binary Cross-Entropy (BCE) prediction loss. A common limitation of CAM is its tendency to activate only the most discriminative object regions. To address this limitation, recent studies have proposed various training strategies, including techniques such as erasing~\cite{wei2017object}, online attention accumulation~\cite{jiang2019integral}, and cross-image semantic mining~\cite{sun2020mining}. Researchers in~\cite{chang2020weakly} suggest leveraging auxiliary tasks to regularize the training objective, such as learning visual words. Contrast pixel and prototype representations~\cite{chen2022self,du2022weakly} to promote the comprehensive activation of object regions. Typically, these methods are built upon the CAM framework, constrained by the drawbacks of CAM. 
The framework of CAM is still limited by its tendency to activate only the most discriminative object regions

\subsection{Vision Transformers.}
Vision Transformer (ViT) has garnered significant success in various vision tasks~\cite{carion2020end,dosovitskiy2020image}. Some recent works also introduce ViT to WSSS, ViT~\cite{dosovitskiy2020image} have emerged as an alternative for generating CAMs~\cite{xu2022multi,ru2022learning}. With models like MCTformer~\cite{xu2022multi}, AFA~\cite{ru2022learning}, although both still rely on the use of CAM. MCTformer leverages ViT's attention mechanism to generate localization maps and employs PSA~\cite{ahn2018learning} for pseudo-mask generation. AFA employs ViT's multi-head self-attention for global insights and an affinity module to propagate pseudo-masks. ViT-PCM~\cite{rossetti2022max} uses patch embeddings and max pooling to infer the probability of pixel-level labels, marking the first instance of employing a framework without relying on CAMs to generate baseline pseudo-masks in WSSS, but it only utilizes max pooling information. However, this approach of employing max pooling might be adversely affected by misclassified patches. In contrast to these approaches, We employ ViT as the backbone, using top-k pooling for the first time to address the limitations caused by misclassified patches, based on a framework without CAM. Indeed, we introduce the PCE to improve intra-class compactness and inter-class separability of patch embeddings to further improve the quality of pseudo labels.

\section{Methodology}
\label{sec:method}

In this section, we will present the overall structure and key components of our proposed approach. We first provide an overview of TKP-PCL method in Sec.~\ref{sec:Overview}.  Subsequently, the proposed top-K pooling layer is described in Sec.~\ref{sec:top-K_Pooling}, aiming to address incorrect predictions dominated by one single patch. 
Finally, in Sec.~\ref{sec:PatchContrast}, the patch contrastive error (PCE) is introduced to enhance the representations of patches to reduce the distance between patches of the same category and increase the distance between patches of different categories in the embedding space.


\subsection{Overall Framework} \label{sec:Overview}
Fig.~\ref{fig:framework} demonstrates the general pipeline and key components of our TKP-PCL. In the first stage, the input image is divided into fixed-size patches, and then an ViT encoder is used to generate the embedding for each patch. Next, HV-BiLSTM~\cite{pmlr-v48-oord16} is employed to refine patch embeddings to better classify each patch. To further improve the performance of patch classifier, top-K pooling and PCE are adopted. The predicted patch-level labels are used as the initial pseudo labels. Finally, conditional random field (CRF)~\cite{krahenbuhl2011efficient} is used to provide the final pseudo labels for segmentation tasks.

\subsection{Patch Embeddings with Top-K Pooling} \label{sec:top-K_Pooling}
The motivation of using top-K pooling is to facilitate the mapping between the patch-level classification and the image-level classification. In the previous work~\cite{rossetti2022max}, global max pooling (GMP) only selects the patch with the highest prediction score for each class, which may map the inaccurate patch-level prediction to the image-level classification. In the experiments, we observed that a few patches are misclassified with a high prediction score. Thus, to enable a more robust projection between the patch-level classification and image-level classification, we average the prediction scores of the $k$ patches with the highest value in each category as the prediction score for image-level classification.

\begin{figure*}[t] 
\vspace{-0.5cm}
\begin{center}
    \includegraphics[width=0.7\linewidth]{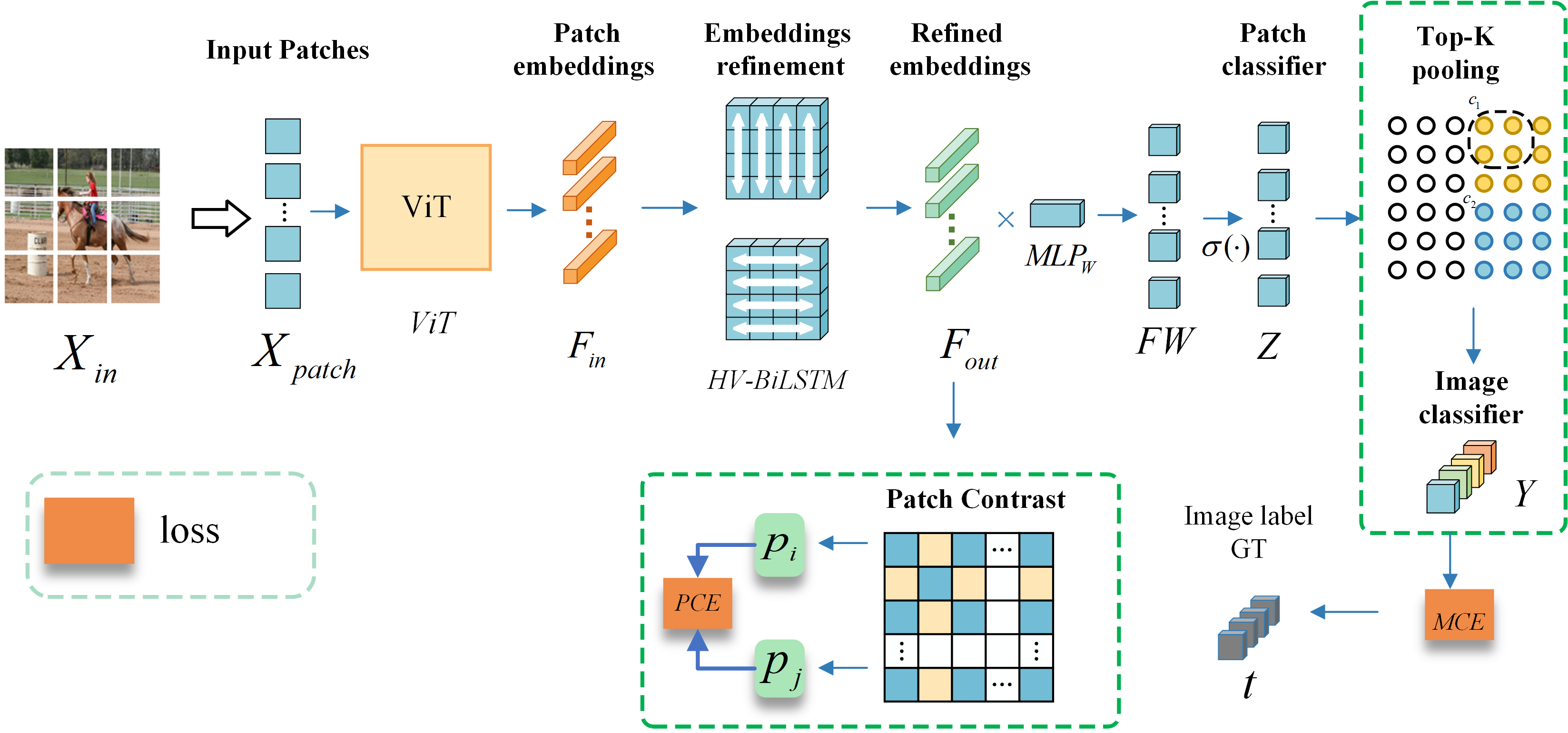}
    \vskip -0.3cm
   \caption{The overall framework of TKP-PCL. Firstly, TKP-PCL utilizes ViT as an encoder to generate patch embeddings. Subsequently, BiLSTM is employed to further refine the patch encoder. Following this, the patch-to-classifier predictions are obtained through MLP and softmax activation. In the top-K pooling module, we establish a mapping from the patch classifier to the image classifier, leveraging image-level GT label for model supervision. Moreover, after the Refined Encoder stage, we introduce a patch contrastive error (PCE) to enhance the incorporation of similarity among patches.}
    \label{fig:TKP-PCL}
\end{center}
\vspace{-0.3cm}
\end{figure*}

The illustration of details about this method is shown in Fig.~\ref{fig:TKP-PCL}. The input image $X_{in}\in \mathbb{R}^{h\times w \times 3}$ is firstly divided into $s$ fixed-size input patches $X_{patch}\in \mathbb{R}^{d\times d \times 3}$, where $s = \frac{hw}{d^{2}}$. Then, the input patches $X_{patch}$ are passed into ViT-based encoder to generate the patch embeddings $F_{in}\in \mathbb{R}^{s\times e}$. Next, HV-BiLSTM is used to enhance the representation of $F_{in}$ and output refined patch embeddings $F_{out}$ with the same size as $F_{in}$. Given the refined patch embedding $F_{out}$, we use a weight $W\in \mathbb{R}^{e\times\mathcal{|C|}}$ and a softmax activation function to predict the class $c\in\mathcal{C}$ of each patch, where $|\mathcal{C}|$ is the total number of classes:
\begin{equation}
\begin{aligned}
    Z=\text{softmax}(F_{out}W).
\end{aligned}
\end{equation}
The output $Z\in \mathbb{R}^{s\times|\mathcal{C}|}$ represents the prediction scores of each class for each patch. To have a robust projection from patch-level prediction to image-level prediction, top-K pooling selects $k$ patches $\{\bar{Z}^{c}_{i=1:k}\}$ with the highest value in each category and computes the average value as the prediction score for image-level classification: $y_c = \frac{1}{k} \sum_{i=1}^{k}\bar{Z}^{c}_{i}$,
where $y_c$ is the projected image-level prediction score of class $c$. Top-K pooling ensures the final image prediction results are not dominated by any misclassified patches, thus further improving the mapping from patch-level prediction to image-level prediction. 
Finally, we minimize the error between the predicted image labels $y_{c}$ and ground-truth labels $t_{c}$ by using the multi-label classification prediction error (MCE):
\begin{equation} 
\begin{aligned}\label{eq:MCE}
\mathcal{L}_{MCE}&=\frac{1}{|\mathcal{C}|}\sum_{c\in\mathcal{C}}{BCE(t_c,y_c)}\\
&=-\frac{1}{\mathcal{|C|}}\sum_{c\in\mathcal{C}}{t_c\log(y_c)+(1-t_c)\log(1-y_c)}.
\end{aligned}
\vspace{-1em}
\end{equation}

\subsection{Patch Contrastive Error} \label{sec:PatchContrast}

To enhance the representation of patch embeddings and thus improve the accuracy of prediction, the patch contrastive error (PCE) is used to narrow the distance between patch embeddings with high prediction scores in a specific category $c$ and also expand the distance between patch embeddings with high scores and the low confidence patch embeddings in the same category. In this work, cosine similarity is used to measure the distance between patch embeddings:
\begin{equation} 
\begin{aligned}\label{eq:cos}
S(F_{out}^{i}, F_{out}^{j}) = \frac{F_{out}^{i} \cdot F_{out}^{j}}{\|F_{out}^{i}\|\|F_{out}^{j}\|},
\end{aligned}
\end{equation}
where a higher similarity value means two patch embeddings are closer to each other, and a lower value indicates a further distance. To represent the similarity more explicitly, we normalize the range of value between 0 and 1 via:
\begin{equation}
    \bar{S}(F_{out}^{i}, F_{out}^{j}) = \frac{1 + {S}(F_{out}^{i}, F_{out}^{j})}{2}.
\end{equation}
As illustrated in Fig.~\ref{fig:TKP-PCL}, $F_{out}$ is used as the input of PCE. Furthermore, the patch prediction score $Z_{i}^{c}$ and a threshold $\epsilon$ are used to determine whether a patch is a high confidence one $\mathcal{P}^{c}_{high} = \{F^{i}_{out}|Z_i^c > \epsilon\}$. Similarly, the patch with the lowest prediction score is considered as a low confidence one $\mathcal{P}^{c}_{low} = \{F^{i}_{out}|Z_i^c < (1 - \epsilon)\}$. Then, the patch contrastive error of a category $c$ can be expressed as:
\begin{equation} 
\begin{aligned}\label{eq:PCE}
\mathcal{L}_{PCE}^{c}&=\frac{1}{N_{pair}^{+}}\sum_{i=1}^{|\mathcal{P}^{c}_{high}|}\sum_{j=1,j\neq i}^{|\mathcal{P}^{c}_{high}|}(1-\bar{S}(F_{high}^{i}, F_{high}^{j}))\\
&+\frac{1}{N_{pair}^{-}}\sum_{m=1}^{|\mathcal{P}^{c}_{high}|}\sum_{n=1}^{|\mathcal{P}^{c}_{low}|} \bar{S}(F_{high}^{m}, F_{low}^{n}),
\end{aligned}
\end{equation}
where $N_{pair}^{+}$ denotes the number of pairs of patches with high confidence, $N_{pair}^{-}$ denotes the number of pairs of high confidence and low confidence patches. $F_{high}$ and $F_{low}$ represent the feature embeddings of high confidence patch $\mathcal{P}^{c}_{high}$ and low confidence patch $\mathcal{P}^{c}_{low}$, respectively. The loss function of TKP-PCL $\mathcal{L}$ can be written as:
\begin{equation}
\mathcal{L} = \mathcal{L}_{MCE} + \alpha\sum_{c\in\mathcal{C}}\mathcal{L}_{PCE}^{c},
\end{equation}
where $\alpha$ is the weight coefficient of $\mathcal{L}_{PCE}$.

\section{Experiments}
\label{sec:Experiments}

In this section, we describe the experimental settings, including dataset, evaluation metrics, and implementation details. We then compare our method with state-of-the-art approaches on PASCAL VOC 2012 dataset~\cite{everingham2010pascal} and MS COCO 2014~\cite{lin2014microsoft}. Finally, ablation studies are performed to validate the effectiveness of crucial components in our proposed method.

\subsection{Experimental Settings}

\textbf{Dataset and Evaluated Metric.} The method is trained and validated on PASCAL VOC 2012~\cite{everingham2010pascal}, which comprises $21$ categories, and MS COCO 2014~\cite{lin2014microsoft}, which comprises $81$ categories, both including an additional background class. The PASCAL VOC 2012 dataset~\cite{everingham2010pascal} is typically augmented with the SBD dataset~\cite{hariharan2011semantic}. To measure the performance of models, mean Intersection-Over-Union (mIoU) is selected as the evaluation metric.

\begin{figure}[h]
\centering
\includegraphics[width=1\linewidth]{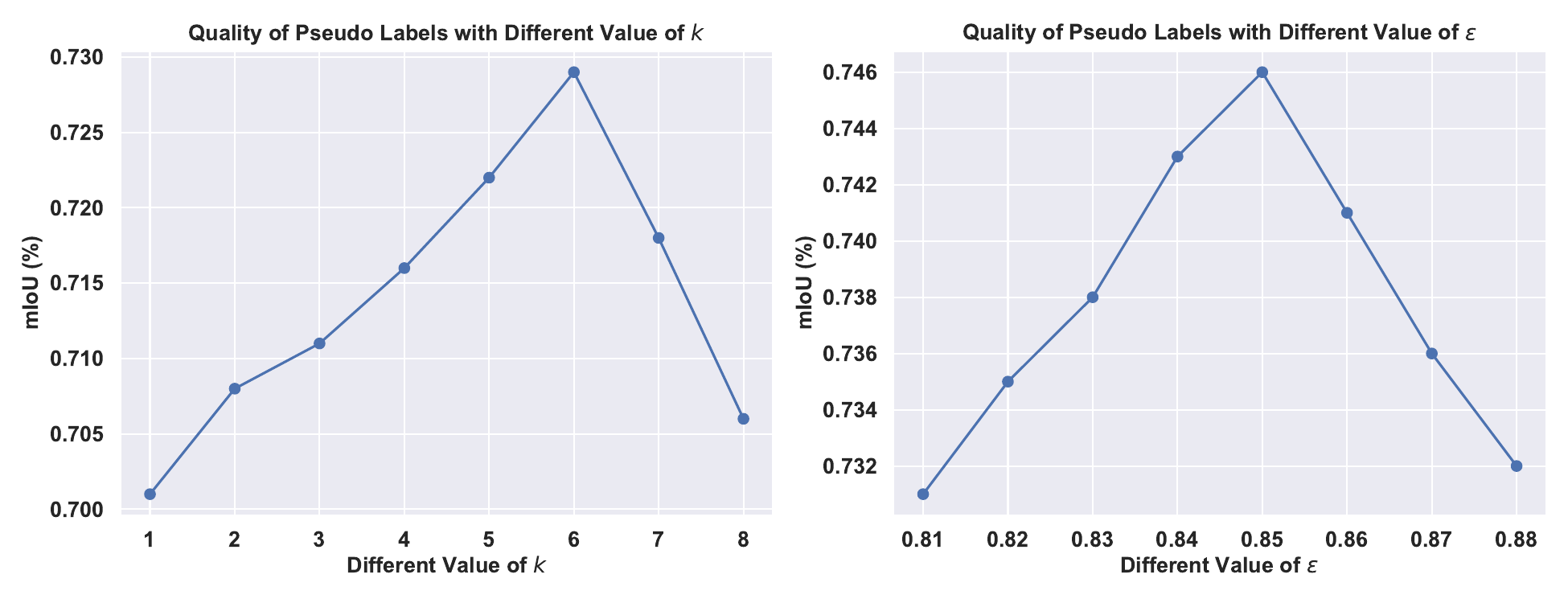}
\vspace{-0.6cm} 
\caption{The performance comparison of selecting different values of $k$ and $\epsilon$.}
\label{fig:diff}
\end{figure}

\noindent\textbf{Implementation Details.} In the experiments, the ViT-B/16 model is used as the backbone for the ViT encoder. During training, input images are first resized into 384×384~\cite{kolesnikov2016seed} and then splitted into 24×24 small patches for the ViT encoder. The model is trained with a batch size of 16 and for a maximum of 50 epochs using two NVIDIA 4090 GPUs. We employ the Adam optimizer and schedule the learning rate as follows: a learning rate of $10^{-3}$ for the initial two epochs, and followed by a learning rate of $10^{-4}$ for the rest of epochs. We set a threshold $\epsilon$ of 0.85 to identify if the patch is a high confidence one for PCE and set $k$ to 6 for top-K pooling. Fig.~\ref{fig:diff} compares the performance by selecting different values for $k$ and $\epsilon$. The weight coefficient $\alpha$ is 0.01. During validation, the pseudo labels of training data are first generated. Then, DeepLab-v2~\cite{chen2018encoder} is chosen to be trained on the training set with pseudo labels and validated on the validation set.


\begin{table}[ht]
\centering
\caption{Results of predicted pseudo masks on VOC train.}
\label{tab:vocbpm}
\begin{adjustbox}{width=0.8\linewidth}
\begin{tabular}{@{}ccccc@{}}
\toprule
method & Pub. & Backbone & mIoU (\%) \\
\midrule

MCTformer~\cite{xu2022multi} & CVPR22 & DeiT-S & 61.7  \\
PPC~\cite{du2022weakly} & CVPR22 & ResNet38 & 61.5  \\
SIPE~\cite{chen2022self} & CVPR22 & ResNet50 & 58.6  \\
AFA~\cite{ru2022learning} & CVPR22 & MiT-B1 & 66.0 \\
ViT-PCM~\cite{dosovitskiy2020image} & ECCV22 & ViT-B/16 & 71.4 \\
ToCo~\cite{ru2023token} & CVPR23 & ViT-B/16 & 72.2 \\
\midrule
\textbf{TKP-PCL (Ours)} &  & ViT-B/16 & \textbf{74.6}  \\
\bottomrule
\end{tabular}
\end{adjustbox}
\end{table}
\vspace{-2em}
\label{sec:conclusion}

\begin{table}[ht]
\centering
\caption{Semantic segmentation performance by using only pseudo masks for training on Pascal VOC 2012 val.}
\label{tab:vocseg}
\begin{adjustbox}{width=0.8\linewidth}
\begin{tabular}{@{}ccccc@{}}

\toprule
Model & Pub. & Backbone & mIoU (\%)\\
\midrule
MCTformer~\cite{xu2022multi} & CVPR22 & DeiT-S & 61.7 \\
PPC~\cite{du2022weakly} & CVPR22 & ResNet38 & 61.5\\
SIPE~\cite{chen2022self} & CVPR22 & ResNet50 & 58.6\\
AFA~\cite{ru2022learning} & CVPR22 & MiT-B1 & 63.8 \\
ViT-PCM~\cite{dosovitskiy2020image} & ECCV22 & ViT-B/16 & 69.3 \\
ToCo~\cite{ru2023token} & CVPR23 & ViT-B/16 & 70.5 \\
SAPC-CS~\cite{sun2023self} & ICME23 & MiT-B1 & 67.5 \\
\midrule
\textbf{TKP-PCL (Ours)} &  & ViT-B/16 & \textbf{72.2 } \\
\bottomrule
\end{tabular}
\end{adjustbox}
\end{table}

\begin{table}[ht]
\centering
\caption{ Semantic segmentation performance by using only
pseudo masks for training on MS-COCO 2014 val set.}
\label{tab:cocoseg}
\begin{adjustbox}{width=0.8\linewidth}
\begin{tabular}{@{}ccccc@{}}

\toprule
Model & Pub. & Backbone & mIoU (\%)\\
\midrule
MCTformer~\cite{xu2022multi} & CVPR22 & Resnet38 & 42.0 \\
SIPE~\cite{chen2022self} & CVPR22 & Resnet38 & 43.6\\
AFA~\cite{ru2022learning} & CVPR22 & MiT-B1 & 38.9 \\
ViT-PCM~\cite{dosovitskiy2020image} & ECCV22 & ViT-B/16 & 45.0 \\
ToCo~\cite{ru2023token} & CVPR23 & ViT-B/16 & 42.3 \\
SAPC-CS~\cite{sun2023self} & ICME23 & MiT-B1 & 39.9 \\
\midrule
\textbf{TKP-PCL (Ours)} &  & ViT-B/16 & \textbf{45.6 } \\
\bottomrule
\end{tabular}
\end{adjustbox}
\end{table}

\subsection{Comparison with State-of-the-arts}
\textbf{Comparison of Pseudo Masks.} Tab.~\ref{tab:vocbpm} shows that our TKP-PCL outperforms other state-of-the-art methods in predicting pseudo masks, achieving an mIoU value of 74.6\% on training data. This is because our proposed method effectively mitigates the limitation of using the prediction score from a single patch to map the patch-level classification into an image-level one, and the proposed patch constrasitve error also further improve the final performance.

\begin{figure}[h!]
\centering
\includegraphics[width=\linewidth]{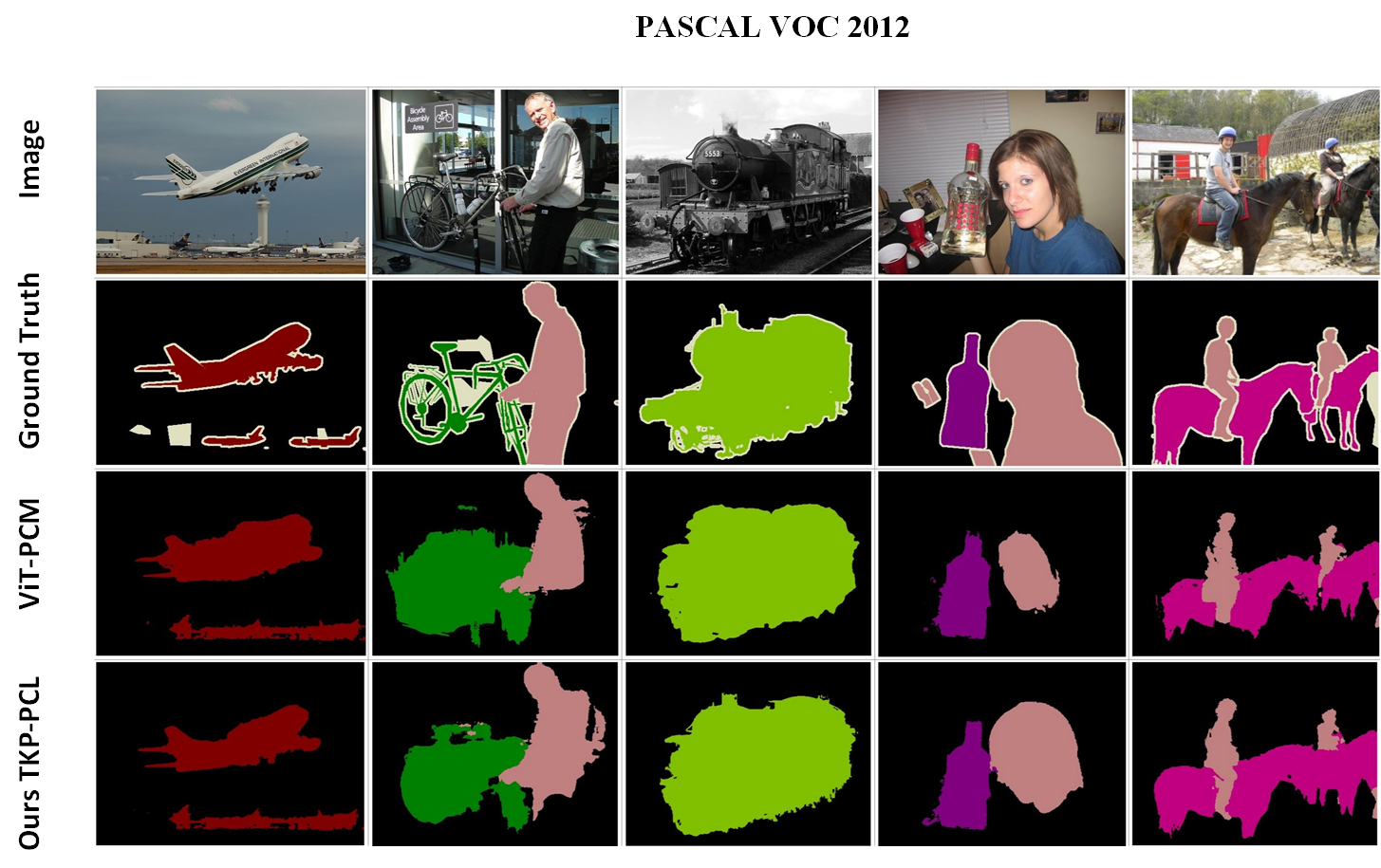}
\vspace{-2em} 
\caption{The comparison of qualitative segmentation results with ViT-PCM~\cite{dosovitskiy2020image}.}
\label{fig:result}
\vspace{-1em} 
\end{figure}

\noindent\textbf{Improvements in Segmentation Results.} We also train the DeepLab-v2~\cite{chen2018encoder} model by using only generated pseudo masks and compare the segmentation results with current state-of-the-art techniques. The results are displayed in Tab.~\ref{tab:vocseg} and Tab.~\ref{tab:cocoseg}, which indicates our method still achieves the best performance. The comparison of qualitative segmentation results are shown in Fig.~\ref{fig:result}.

\begin{table}[h!]
\centering
\caption{Ablation studies on top-K pooling and patch contrastive error (PCE).}
\label{tab:ablation}
\begin{adjustbox}{width=\linewidth}
\begin{tabular}{ccccccc}
\toprule
Backbone & Avg-Pooling & Max-Pooling & top-K & PCE & Mask mIoU (\%) & Seg mIoU (\%)\\
\midrule
ViT-B/16 & \checkmark & &  &  & 60.5\% & 59.6\% \\
ViT-B/16 & & \checkmark &  &  & 70.1\% & 68.6\% \\
ViT-B/16 & &  & \checkmark & & 72.9\% & 69.8\% \\
ViT-B/16 & &  & \checkmark & \checkmark & 74.6\% & 72.2\% \\
\bottomrule
\end{tabular}
\end{adjustbox}
\vspace{-1.8em}
\end{table}

\subsection{Ablation Studies}
Ablation studies are conducted to verify the effectiveness of two key components of our proposed method: top-K pooling and patch contrastive error (PCE). As shown in Tab.~\ref{tab:ablation}, top-K pooling improves the quality of pseudo mask by 2.8\% mIoU and 12.4\% mIoU over max pooling and average pooling. It also facilitates the performance of segmentation. In addition, by combining the proposed PCE, the quality of pesudo masks and the performance of segmentation task are further improved by 1.7\% and 2.4\% in mIoU, respectively. Thus, these two components help our approach outperform previous ViT-based methods significantly.

\section{Conclusion}

In this work, we propose TKP-PCL approach without using CAM for weakly supervised semantic segmentation. Unlike previous methods, TKP-PCL selects top-K patches for mapping patch-level classification into image-level classification, thus alleviate the issue of potential misclassified patches. In addition, patch contrastive error (PCE) is proposed to further enhance the feature embeddings of patches. In the same class, PCE aims to decrease the distance between patch embeddings with high confidence and increase the distance between embeddings with high confidence and embeddings with low confidence. By combining these two components, our method achieves state-of-the-art results in weakly supervised semantic segmentation tasks using only image-level labels.


\bibliographystyle{IEEEbib}
\bibliography{refs}

\end{document}